\documentclass[letterpaper, 10 pt, conference]{ieeeconf}  % Comment this line out if you need a4paper

\IEEEoverridecommandlockouts                              % This command is only needed if 
\usepackage{amssymb}
\usepackage{amsmath}

\usepackage{url}
\usepackage{hyperref}

\usepackage{bm}
\usepackage{upgreek}

\usepackage{cite}
\usepackage[dvips]{graphicx}
\usepackage{array}
\usepackage{subcaption}
\usepackage{float}
\usepackage{graphicx}
\usepackage{multirow}
\usepackage[export]{adjustbox}
\usepackage{color, soul}
\sethlcolor{white}
\makeatletter
\def\SOUL@hlpreamble{%
	\setul{\dp\strutbox}{\dimexpr\ht\strutbox+\dp\strutbox\relax}%
	\let\SOUL@stcolor\SOUL@hlcolor
	\SOUL@stpreamble
}
\makeatother
\makeatother

\usepackage[skip=2pt]{caption} % example skip set to 2pt
\usepackage[font=small,labelfont=bf]{caption}
\definecolor{mygreen}{RGB}{28,172,0} % color values Red, Green, Blue
\definecolor{mylilas}{RGB}{170,55,241}
\usepackage{color, colortbl}
\definecolor{Gray}{gray}{0.9}
\definecolor{LightCyan}{rgb}{0.88,1,1}
\usepackage{graphicx}
\usepackage{color, soul}
\sethlcolor{white}
\makeatletter
\def\SOUL@hlpreamble{%
	\setul{\dp\strutbox}{\dimexpr\ht\strutbox+\dp\strutbox\relax}%
	\let\SOUL@stcolor\SOUL@hlcolor
	\SOUL@stpreamble
}
\makeatother
\title{\LARGE \bf
WaspL:	Design of a Reconfigurable Logistic Robot for Hospital Settings}

\author{ Yuyao Shi, A. A. Hayat, S. Vinu and M. R. Elara% <-this % stops a space
	\thanks{*This work is supported by the National Robotics Research and Development Programme Office (NR2PO) Singapore}% <-this % stops a space
	\thanks{Shi Yuyao is research officer at SUTD in ROAR Lab,
		{\tt\small yuyao\_shi@sutd.edu.sg}}%
	\thanks{Abdullah Aamir Hayat is research fellow at SUTD in ROAR Lab, 
		{\tt\small abdullahaamir@sutd.edu.sg}}%{\tt\small rajeshelara@sutd.edu.sg}}%	
    \thanks{Vinu is a master student at SUTD in ROAR lab, {\tt\small vnu.619@gmail.com}}
	\thanks{Mohan Rajesh Elara is assistant professor at Engineering Product Development Pillar, Singapore University of Technology and Design (SUTD), 
		{\tt\small rajeshelara@sutd.edu.sg}}%	
}

\begin{document}

\maketitle
\thispagestyle{empty}
\pagestyle{empty}

%%%%%%%%%%%%%%%%%%%%%%%%%%%%%%%%%%%%%%%%%%%%%%%%%%%%%%%%%%%%%%%%%%%%%%%%%%%%%%%%
%%%%%%%%%%%%%%%%%%%%%%%%%%%%%%%%%%%%%%%%%%%%%%%%%%%%%%%%%%%%%%%%%%%%%%%%%%%%%%%%
\begin{abstract}
Healthcare poses diverse logistic requirements, which resulted in the deployment of several distinctly designed robots within a hospital setting. Each robot comes with its overheads in the form of, namely,  none/limited scaling, dedicated charging stations, programming interface, closed architecture, training requirements, etc. This paper reports on developing a reconfigurable logistic robot named \textit{WaspL}. The design of \textit{WaspL} caters to the requirement of high mobility, open robotic operating system architecture, multi-functionality, and evolvability features. It fulfills multiple logistics modes, like towing, lifting heavy payloads, forklifting low ground clearance objects, nesting of two \textit{WaspL}, etc., fulfilling different applications required in hospital settings.  The design requirements, mechanical layout, and system architecture are discussed in detail. The finite element modeling, attribute-based comparison with other standard robots, are presented along with experimental results supporting the \textit{WaspL} design capabilities.
\end{abstract}

%%%%%%%%%%%%%%%%%%%%%%%%%%%%%%%%%%%%%%%%%%%%%%%%%%%%%%%%%%%%%%%%%%%%%%%%%%%%%%%%
\section{Introduction}
\label{sec:introduction}
Hospitals are an integral part of modern society and deliver their services 24/7. With the increase in the elderly population, demographic factors, and cost control measures, healthcare institutions continuously improve their operations and productivity. Moreover, the post-pandemic scenario reduced physical distancing among humans and increased applications like disinfection of high touch areas, security and inspection, inventory management, logistics in healthcare, etc. By 2026, the automated guided logistic market is expected to increase double-fold, reaching 30 Billion \cite{IQLogistic} with an approximate share of over 28\% in medical robots \cite{report}. The COVID-19 is pushing for innovative solutions to be applied in health care. In particualr, robots are showing huge potentials \cite{tamantini2021robotic}.  The changing demands paved the way for a different design of robots deployments, which comes with its overheads in form of additional requirement of space, charging station, programming interface, training, etc.   Moreover, deploying different robots in limited healthcare settings is not a viable solution. We explore shared benefits with the design requirements inside hospital settings by designing a reconfigurable robot with scalable logistic features named \textit{WaspL}.

The logistics or transportation practices can be broadly classified into three categories, namely,  a) Lifting a payload on top, b) Forklifting a payload with lower ground clearance, and c) Towing a trolley load with a wheelbase. The requirements and applications of autonomous mobile robotics (AMR) in hospital settings are surveyed in \cite{kriegel2021requirements}. Several commercial robots are designed to solve logistic requirements for nursing, pharmacy, medical tests, linens, central supply, etc., with just using one of these modes depending upon the type of payload. Commercial products using lifting mode, such as Swisslog \cite{SwissLog}, TUG \cite{Tug}, MiR \cite{mir_robot}, are successfully installed in hospital. The only constraint is that the robot must go under the cart, which should be high enough to accommodate the robot. It mostly results in dedicated carts for each robot.  Moreover, the payload to weight ratio of these robots are generally low, with mobility usually provided by differential wheels.

Automated guided vehicles (AGV) with towing and forklifting mode are widely reported in factories and warehouse environments. The towing mode using a developed autonomous mobile robots for the warehouses and industry usage is reported in \cite{fadzil2013development} with its performance test results to follow the desired commanded velocity. Autonomous towing of large loads with rigid coupling using the developed robot OzTug can move on guided track in a manufacturing environment is reported in \cite{ horan2011oztug}.  However, the towing mode for logistics in hospitals is not reported, especially with no changes required to be incorporated in the existing trolley in a hospital settings. 

The forklift mode with a dual cantilever arm extending to carry the heavy load using AGVs as pallet vehicles is standard in warehouses. The forklift for pallet handling using a robotic system named Robolift is proposed in \cite{ garibott1996robolift}. The mechanism for the forklift design is described in \cite{ wang2010innovative}. However, the forklift mode in hospital settings is not reported. We have used the concept of forklift using the single cantilever extending out to carry small trolley or lighter and suitable payload in hospital settings. The mechanism for fork lifting in \textit{WaspL} is coupled with the central payload lifting, which helps keep the size of the robot limited to its footprint.

Machines that can change their morphologies as per the functional requirement or adapt to the environment need to be reconfigurable. The principle of reconfiguration in designing a robot or product helps in, multiability, evolvability, and survivability \cite{ siddiqi2008modeling }. Reconfiguration in robotics are further classified as inter-, intra- and nested reconfiguration \cite{ tan2020framework}.  

%Cite inter, intra, and nested paper and then talk about Wasp feature!

%\textcolor{cyan}{[Para 3] Reconfigurable logistic robot (Existing literature and limitation] OR Highlight the work of TRL and then put an arugument that in different mode it is essential to test the path tracking which will alow it to be safely used in the hospital settings} \\

%[Para 4] Design objectives of the designed robot \textit{WaspL} \\
%Where is the actual contribution of the paper and what is novel? 

Following objectives are set for the present work with the contributions detailed in the paper:
\begin{itemize}
	\item Development of the cross-functional logistic robot using the principle of collaborative engineering.
	\item 	Illustrate the three modes of logistics in a single platform named as \textit{WaspL} highlighting hosptial setting needs.
	\item 	Showcase the scaling feature of the base platform \textit{WaspL} by depicting different use-cases.
	\item   The teleoperated path following tests are performed with different logistic modes.
	
\end{itemize}
The rest of this paper is organized as follows. Section \ref{sec:waspdesign} explains on the design principle and mechanical features of \textit{WaspL} in detail. The design of \textit{WaspL} is then used to compare with other logistic robots. Section \ref{sec:use_cases} describes the different mode of logistics and other extended use cases that has been tested with the \textit{WaspL}. Section~\ref{sec:experi} explains the mobility of the \textit{WaspL} while transporting different types of payloads in suitable mode of logistic. Finally, Section~\ref{sec:conclusion} concludes the paper with future work.
%%%%%%%%%%%%%%%%%%%%%%%%%%%%%%%%%%%%%%%%%%%%%%%%%%%%%%%%%%%%%%%%%%%%%%%%%%%%%%%%
%%%%%%%%%%%%%%%%%%%%%%%%%%%%%%%%%%%%%%%%%%%%%%%%%%%%%%%%%%%%%%%%%%%%%%%%%%%%%%%%
\section{W{asp}L Design}\label{sec:waspdesign}

This section presents the essential design requirements obtained from the Changi General Hospital (CGH) by engaging with the hospital staff and management during the multiple site visits. The nature of multi-functional requirements for creating the robot resulted in adopting transformation design principles \cite{singh2009innovations, weaver2010transformation} as enablers for designing the reconfigurable robot are presented. The overall mechanical design with the reconfiguring mode of logistics and transportation as per the requirements are presented.

\begin{figure}
	\centering
	\includegraphics[width=0.99\linewidth]{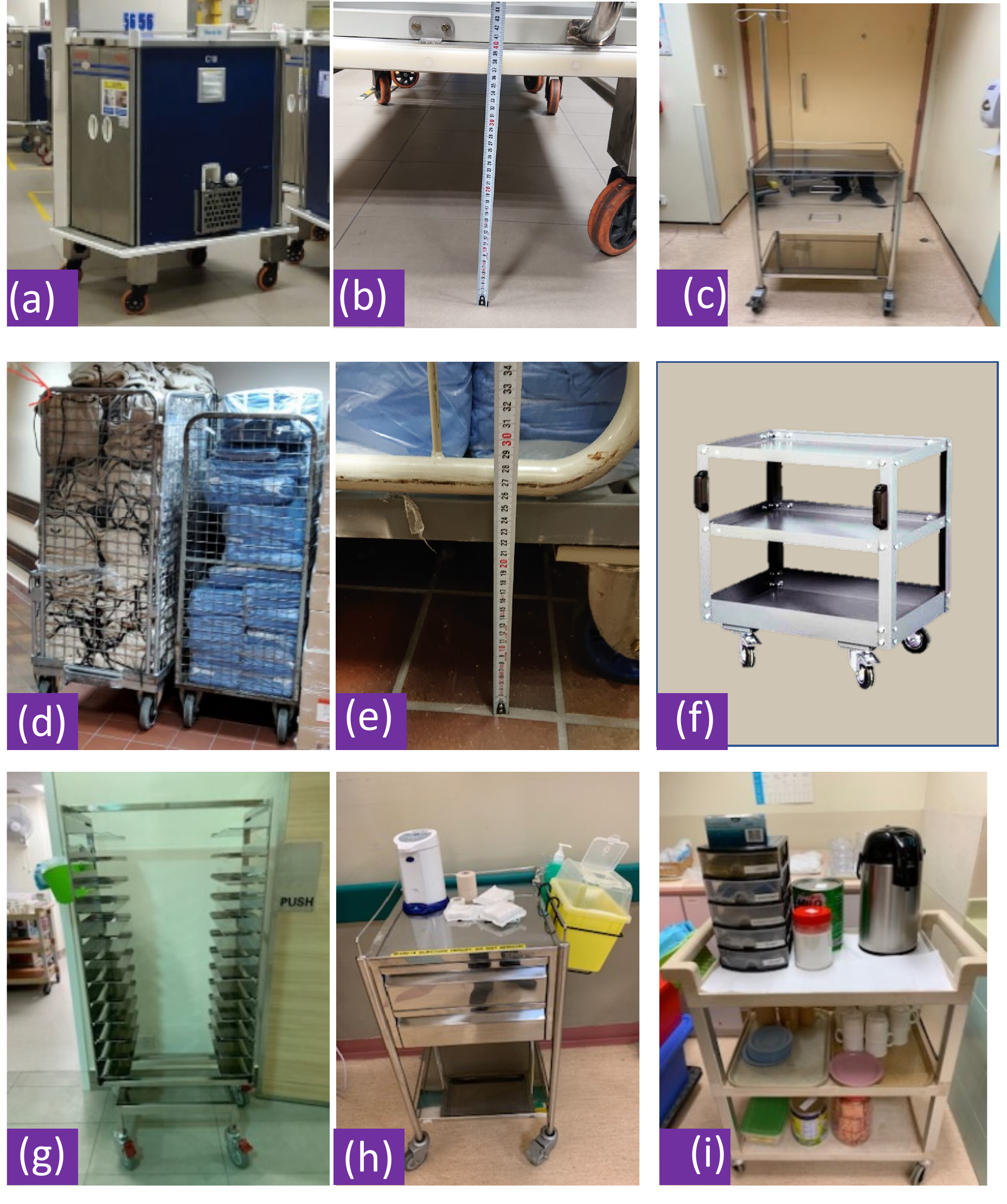}
	\caption{Trolleys inside a hospital settings}
	\label{fig:designrrssinglecolumn13}
\end{figure}

%%%%%%%%%%%%%%%%%%%%%%%%%%%%%%%%%%%%%%%%%%%%%%%%%%%%%%%%%%%%%%%%%%%%%%%%%%%%%%%%
\subsection{Design principles}
\label{sec:DP}
The notion of the logistic robots inside the hospital settings is to provide the transportation of different kind of trolleys used for different purposes, like, pharmacy, linen, food, etc. Fig.~\ref{fig:designrrssinglecolumn13} shows some of the trolleys used inside the hospital settings. Fig.~\ref{fig:designrrssinglecolumn13}a and b shows the food trolley with carrying capacity of upto 200 Kgs and the ground clearance (GC) of about 36 cms respectively. Also Fig.~\ref{fig:designrrssinglecolumn13}g is a food tray trolley which can rack multiple trays has GC of 35 cms.  Fig.~\ref{fig:designrrssinglecolumn13}(d and e) shows the fully loaded linen trolley weighing upto 300 Kgs and GC of 22 cms. Fig.~\ref{fig:designrrssinglecolumn13}(c, f, h and i) are light weight transpiration trolleys with a GC less than 15 cms. The observations and feedback gained on the existing transportation aspects from robot and users aspects are: a) Three different type of robots are used to handle the logistic tasks, b) The mobility of robot deployed are using differential wheels, c) Each robot has its dedicated charging units which adds the space requirement, d) The dimension of the robot restricts more than one occupancy of robot in the lifts, e) The scalability function is limited, f) The system architecture is closed and different UI/UX are used for different robots, g) The  noise level for the robot used in transporting food gets more than 65 dB (whereas, Singapore's National Environmental Agency (NEA) guidelines for maximum permissible noise in day time is 60 dB for hospitals), which restricts its navigation inside wards, f) Light weight trolleys have all four swivel caster wheels whereas the heavy trolleys have two fixed and two swivel caster wheels, g) The wide range of weight and variations in height of trolley for transportation which has to be repetitively moved by human results in musculoskeleton issues and are reported by the hospital staffs. These factors resulted in the following design features:
\begin{itemize}
	\item The platform should have the ability to transport most of the trolleys used in the hospital settings while keeping its footprint minimum. 
	\item The  robot should have high maneuverability and small turning radius to navigate through narrow corridors with less obstruction to human flow.
	\item The platform should be scalable. Scalability here refers to the other modules, like, disinfection, wall cleaning, surveillance, and patient assisting with the modular attachment of essential features.
	\item The maximum payload as per the current usage and requirement should be in a range of 350-400 Kgs.
	\item The noise level should stick to the standards provided by National Environmental Agency (NEA), i.e., typically less the 60 dB for hospital setting.
	\item Final deployable version should adhere to the detailed standards, such as essential safety measures, in terms of speed, e-stop, etc. The guidelines are presently being developed. 	 
\end{itemize}  
To satisfy the multi-functional specification by the robot, the transformation principle and facilitators were useful in selecting the design constructs. Twenty facilitators and three principles for transformation \cite{weaver2010transformation} helped decide the design constructs as per the requirement posed. The mechanical design is detailed in the next section, briefly commenting on the transformation principles and facilitators used in \textit{WaspL}.
%%%%%%%%%%%%%%%%%%%%%%%%%%%%%%%%%%%%%%%%%%%%%%%%%%%%%%%%%%%%%%%%%%%%%%%%%%%%%%%%
\subsection{Mechanical layout}
\label{subsec:mech_design}

\begin{figure}[]
	\centering
	\includegraphics[width=3.4in]{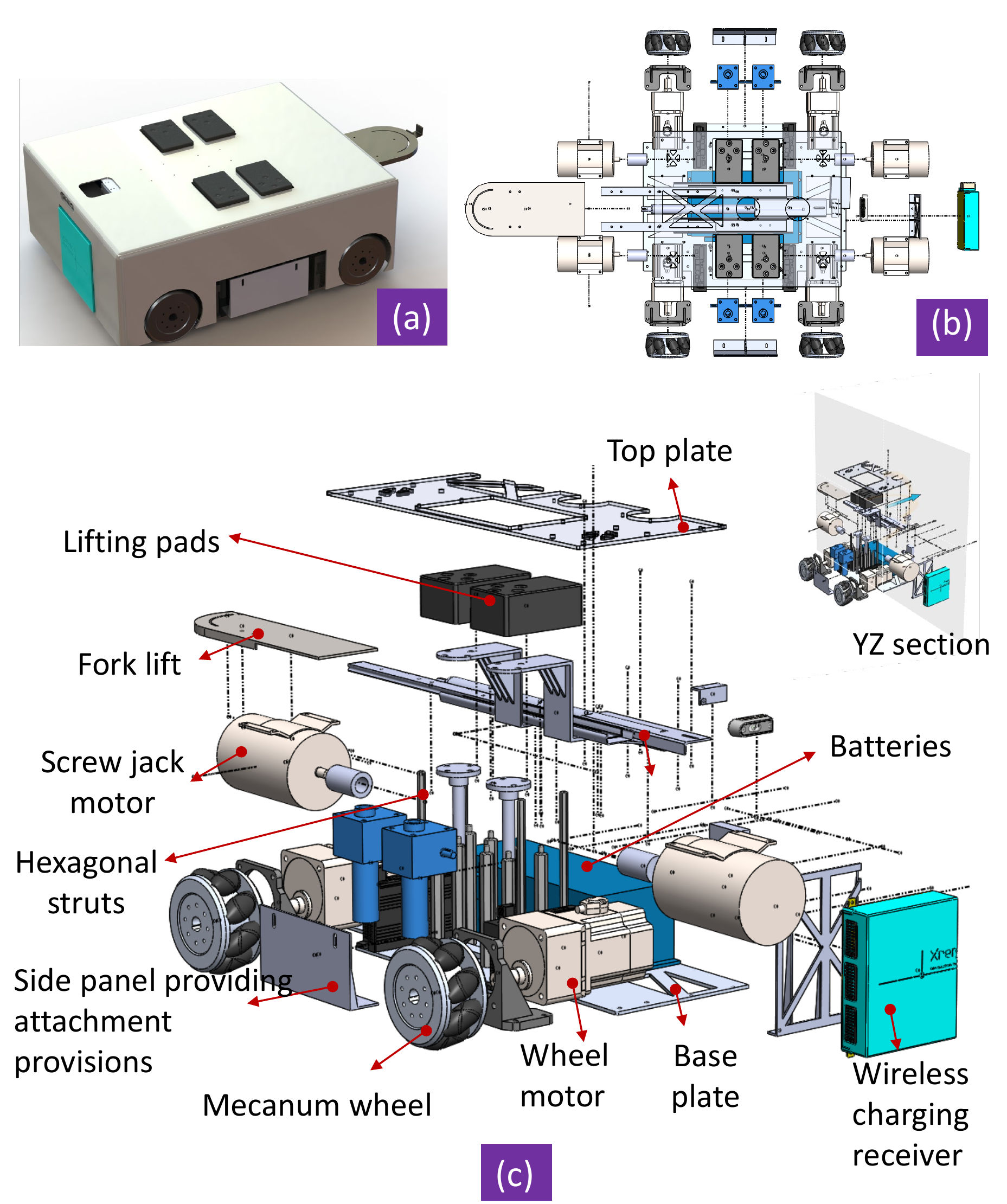}
	\caption{Exploded view of WASP platform}
	\label{fig:archiwaspl}
\end{figure}

In order to fulfill the requirements listed above, the mechanical design is made to perform three different types of payload transportation. As the dimensions, features and weights of each trolley varies a lot from one to another, it is hard to design a unitary transportation solution to fit all the cases while restricting the overall dimension to be under the requirement. 
\begin{figure}
	\centering
	\includegraphics[width=0.9\linewidth]{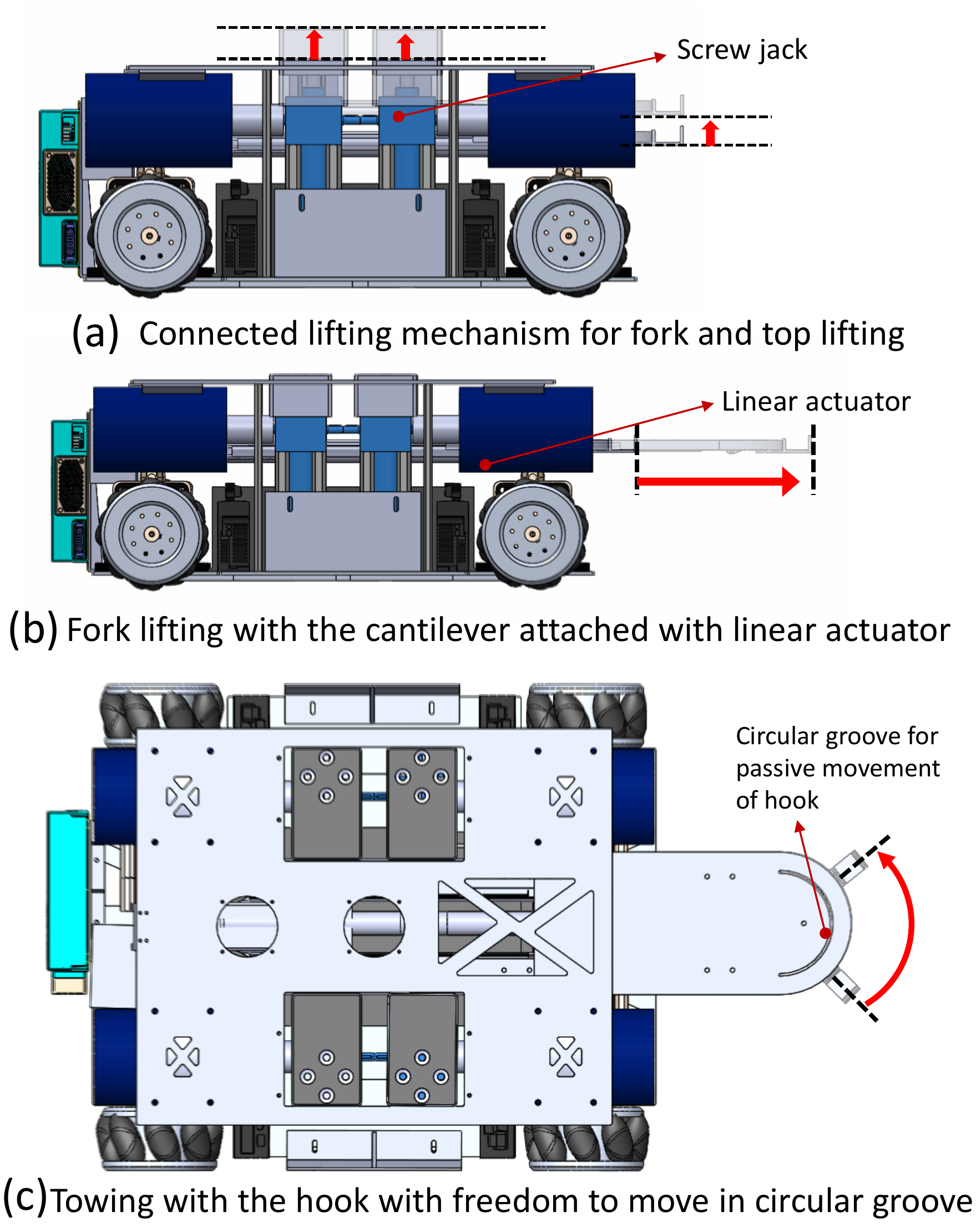}
	\caption{Three modes of logistics using \textit{WaspL}}
	\label{fig:logimode}
\end{figure}

Therefore, the proposed platform is equipped with three transportation methods for different payloads. The design construct is an example of ``common core structure" \cite{weaver2010transformation} for \textit{WaspL}.  As shown in Fig. \ref{fig:logimode}a, is located on the top of the robot which is used to lift up heavy trolleys with higher ground clearance, while the cantilever presented in Fig. \ref{fig:logimode}b is designed for lifting light trolleys with lower ground clearance. As for the heavy trolleys with lower ground clearance, the hook located at the end of the cantilever, showed in Fig. \ref{fig:logimode}c, can be used to tow the trolley. By combining the three mechanisms of shifting payloads together, the robot is capable of transporting majority of the trolleys located in the hospital.

The overall exploded view of this robot is shown in Fig. \ref{fig:archiwaspl}. It consists of four Mecanum wheels driven by 48V brushless DC motors, allowing the platform's maximum speed to be 0.6m/s. Four DC motors power four screw jacks of 0.5kN to lift up the payload. These screjack are self-locking and without power, they can keep the load lifted. Due to the ground clearance variety of the hospital trolleys, the screw jacks are connected to two different lifting modules. The first lifting module is on the top of the robot, which is used for a trolley with 320mm ground clearance and above. The second one is a single forklift module located at the back of the robot, which is designed for lifting or towing payloads with lower ground clearance. To minimize the robot's overall dimension and the number of actuators, the two lifting modules are interconnected using ``shared power transmission” as facilitators through a rigid connector. Hence, only one set of 4 screw jacks is used to lift the lifting pads on top and single forklift . On top of that, a linear actuator is used on the horizontal configuration to extrude the forklift from the back of the robot. The single forklift mode is inspired from the facilitator, namely, the ``telescope” . 

\begin{figure}
    \centering
    \includegraphics[width = 3.3in]{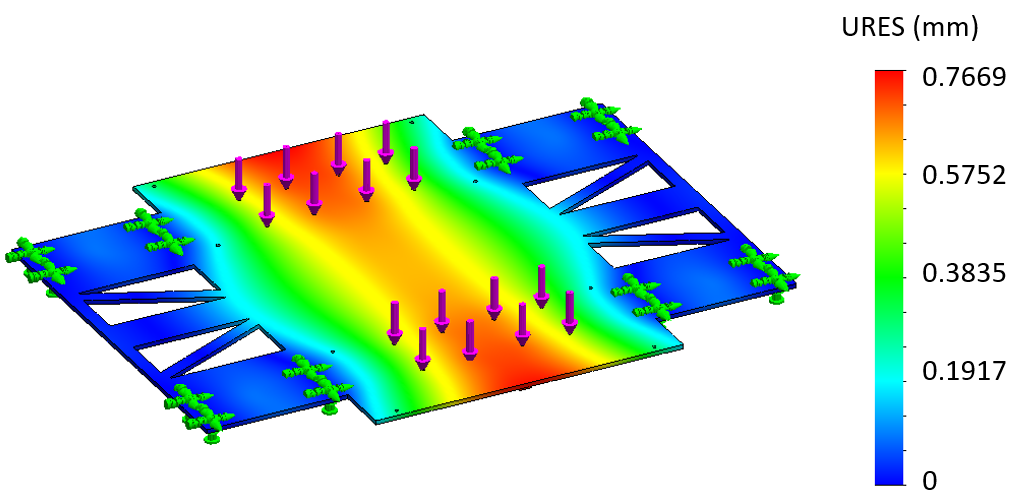}
    \caption{Displacement analysis using FEA with Von-Mises Critera}
    \label{fig:FEA}
\end{figure}

As the trolley is designed for heavy payloads, Finite Element Analysis (FEA) is used to validate the base plate's design, which bears the load's. The structure is designed with ribs at the bottom to increase the efficiency of carrying a payload, since under the same power of motor, the lighter the robot is, the heavier the payload can be achieved. As shown in Fig. \ref{fig:FEA}, the result of FEA demonstrates that the deformation with 400kg payload is only about 0.6mm. The technical specifications for the designed robot are listed in Table~\ref{tab:dimension}.

The wireless charging unit is placed at the center back of the platform. The chraging is universal and the powering unit on the mobile base enables it to position as per the space. The wireless charging is the solution taken from Xnergy \cite{Xnergy}, with its fast charging capability to fully charge the batteries in 50 minutes. The power source is provided by 16 cells Lithium-ion batteries of 48 volts. The use of wireless charging method helps in making the charging unit contactless and lowers the need for precise docking methods using mechanical connectors.
\begin{table}[t]
	\caption{Technical specification for \textit{WaspL} robot}
	\label{tab:specs}
	\begin{tabular}{l|l}
		Item & Specification \\
		\hline
		\rowcolor[rgb]{ .851,  .851,  .851}		Mobility ~~~~~~~~~~~~~~~~~~	& Holonomic	~~~~~~~~~~~~~~~~~~~~~~~~~~~~~~~~~~~ \\
		Mass of \textit{WaspL} 	& 120 Kgs	 \\
		\rowcolor[rgb]{ .851,  .851,  .851} Dimension(mm)	& 625 $\times$780$\times$310 (including cover) \\
		Payload $|$ Lifting	& 400 Kg  \\	
		\rowcolor[rgb]{ .851,  .851,  .851}	Payload $|$ Forklifting	&  90 Kg \\
		Payload $|$ Towing	& 300 Kg		 \\
		%		Motor driver	& Roboclaw \\	
		%		Processing unit  	& Arduino Mega \\
		\rowcolor[rgb]{ .851,  .851,  .851}	Noise level 	& $<$ 58 dB \\
		\hline
	\end{tabular}
\end{table}

%%%%%%%%%%%%%%%%%%%%%%%%%%%%%%%%%%%%%%%%%%%%%%%%%%%%%%%%%5
\subsection{Attribute based comparison}

In this subsection, the comparison between \textit{WaspL} platform and the existing logistic robots is performed based on the attributes extracted from the case study. The attributes used to validate the design are 1) Maximum payload capacity 2) Footprint dimension of the robot 3) Minimum and maximum lifting height of the robot and 4) Maneuverability. Noise level is not selected to be a pertinent attributes as not all logistic robots presented are specifically designed for hospital settings. Table. \ref{tab:dimension} compares these attributes between \textit{WaspL} and existing products. 

\begin{table}[htbp]
	\centering
	\caption{Specification of the logistic robots for hospital settings}
	\begin{tabular}{l|rrrrrr}
		\rowcolor[rgb]{ .851,  .851,  .851} Name  & {P} & {L} & {W} & {HMn} & {HMx} & M \\
		\hline
		\textit{WaspL} & 400   & 740   & 625   & 206     & 365     & 3 \\
		MSA AGVS \cite{MSA_AGV} & 500   & 1750  & 652   & 345     & 385     & 2 \\
		Elfin \cite{Elfin_OKGAV} & 400   & 900   & 700   & 310     & 370     & 2 \\
		Uwant \cite{Uwant} & 400    & 900   & 500   & 320     & 365     & 2 \\
		OmniTurtle \cite{OmniTurtle} & 450   & 1500  & 1000  & 340     & 400     & 3 \\
		Mircolomay \cite{MicroLomy} & 500   & 910   & 700   & 360    & 420    & 2 \\
		Kuka-1500 \cite{KUKA_Robot} & 935   & 2000  & 800   & 470     & 670     & 3 \\
		\hline
		\multicolumn{7}{c}{%
			\begin{minipage}{3in}%
				\small 
				P: Payload, L: length, W: width, HMn: Height Minimum, HMx: Height Maximum, M: Maneuverability \\				
			\end{minipage}%
		}\\ 
	\end{tabular}%
	\label{tab:dimension}%
\end{table}%

\begin{figure}
	\centering
	\includegraphics[width=0.99\linewidth]{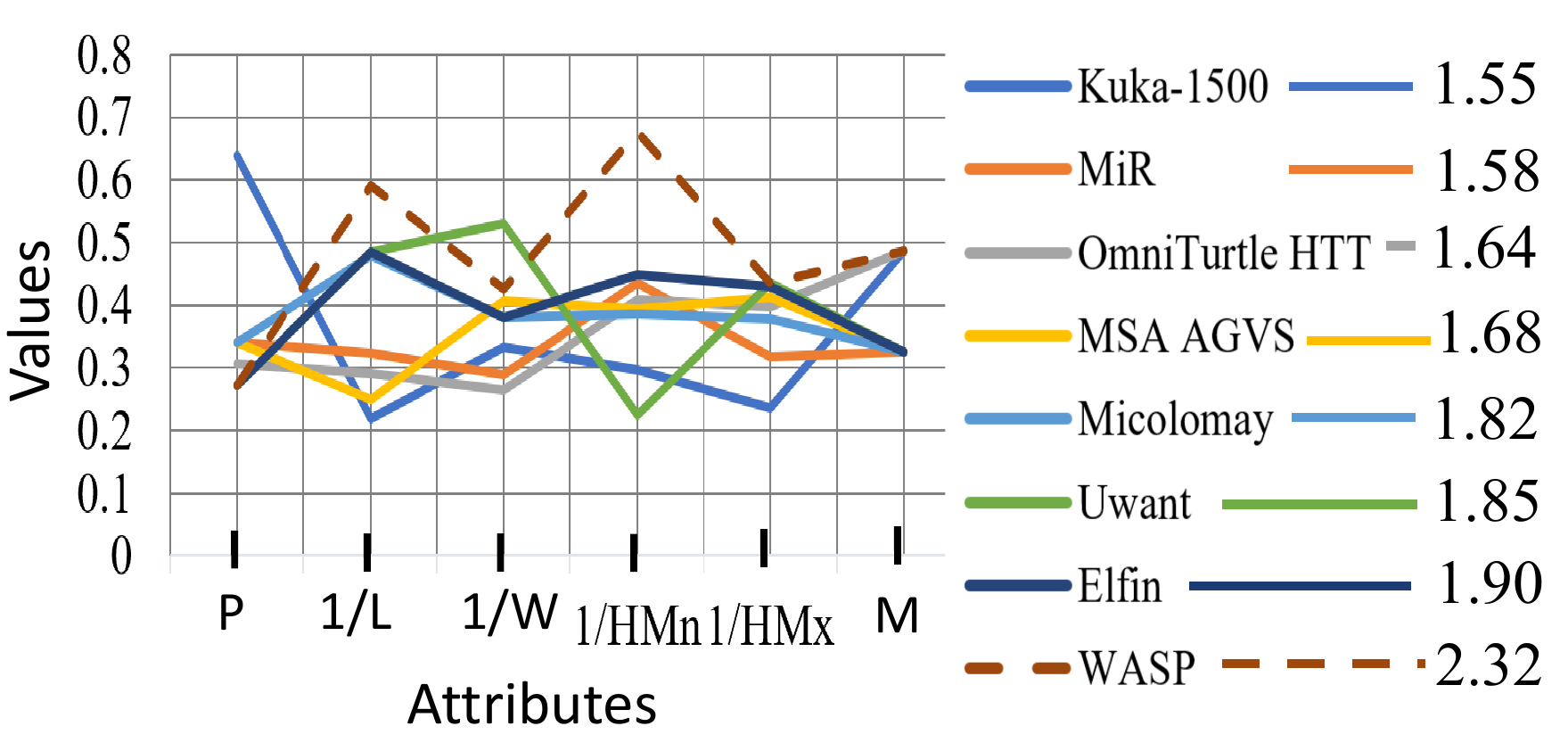}
	\caption{Line graph for comparison of \textit{WaspL} with similar class of robots and the numerical value as scores indicated against each robots next to legends of the graph}
	\label{fig:plotscomparision}
\end{figure}

To better visualize the difference between the \textit{WaspL} platform and existing logistic robots, the data in Table \ref{tab:dimension} is normalized between 0 to 1 and plotted in a line graph shown in Fig. \ref{fig:plotscomparision}, with the normalized value of the attributes on vertical axis and the corresponding attribute on the abscissa. The score was be calculated based on the area under each line. As the length, width and minimum lifting height are attributes, of which the values are preferred to be small, their reciprocals are used in the plot. From the plot, the area under the lines for each platform is calculated and displayed next to the name of the platform in Fig. \ref{fig:plotscomparision}. It shows that the intra-reconfigurability of the \textit{WaspL} platform makes it a more universal logistic robot. In other words, for payloads upto 400kg, \textit{WaspL} platform, in general, has smaller dimension and larger range of lifting methods different payloads.

%Design principal
%Mechanical Architecture
%Reconfigurable payloads
\label{sec:dyn}
\section{Application Scenarios}\label{sec:use_cases}
\begin{figure}[!b]
	\centering
	\includegraphics[width=0.9\linewidth]{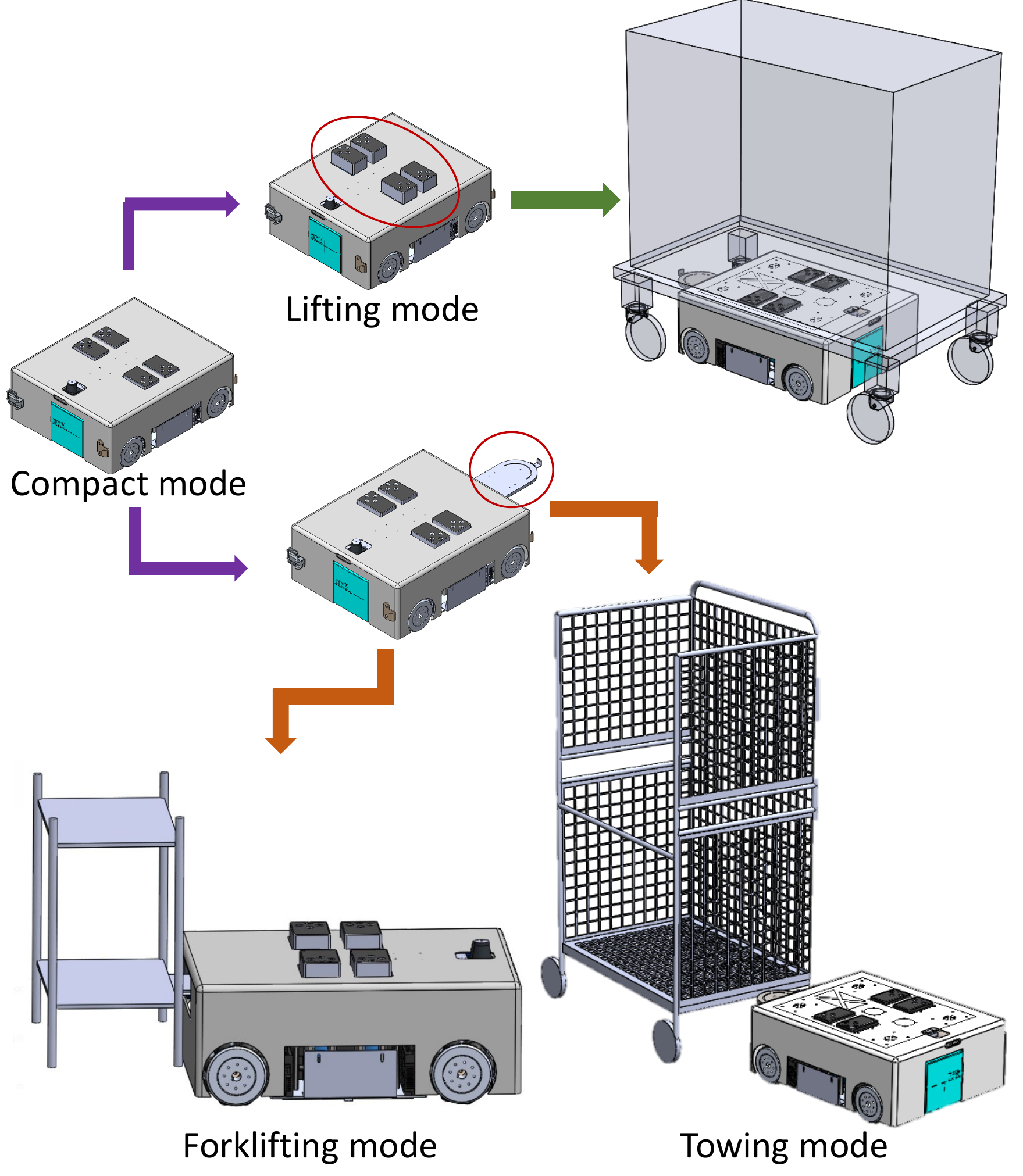}
	\caption{Intra-reconfigurability in the logist mode for lifting, towing and forklifting.}
	\label{fig:modesusecase}
\end{figure}
\begin{figure}
	\centering
	\includegraphics[width=0.99\linewidth]{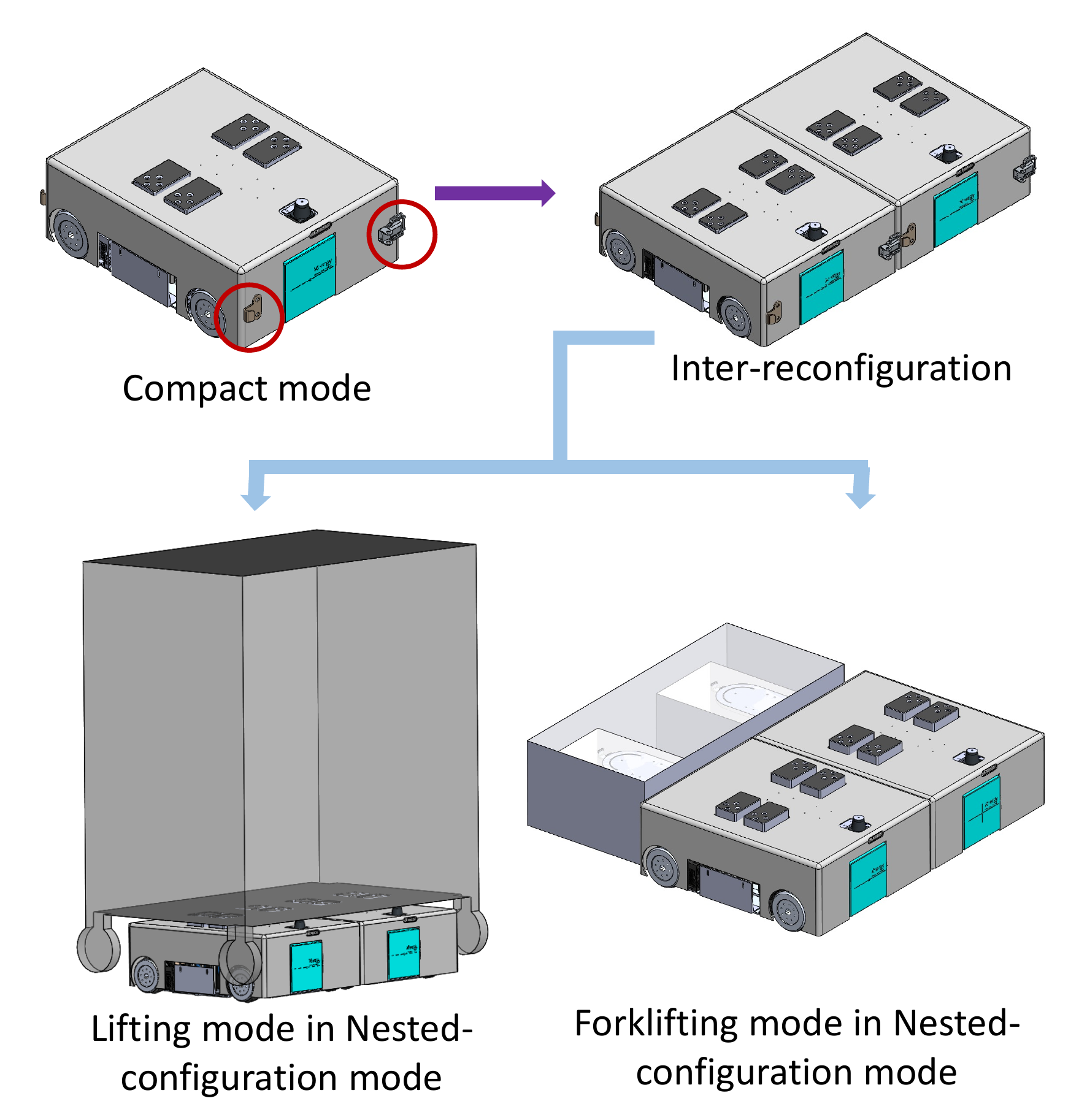}
	\caption{Inter- and Nested-reconfiguration case with \textit{WaspL} }
	\label{fig:modesnested}
\end{figure}
\begin{figure}[!b]
	\centering
	\includegraphics[width=0.99\linewidth]{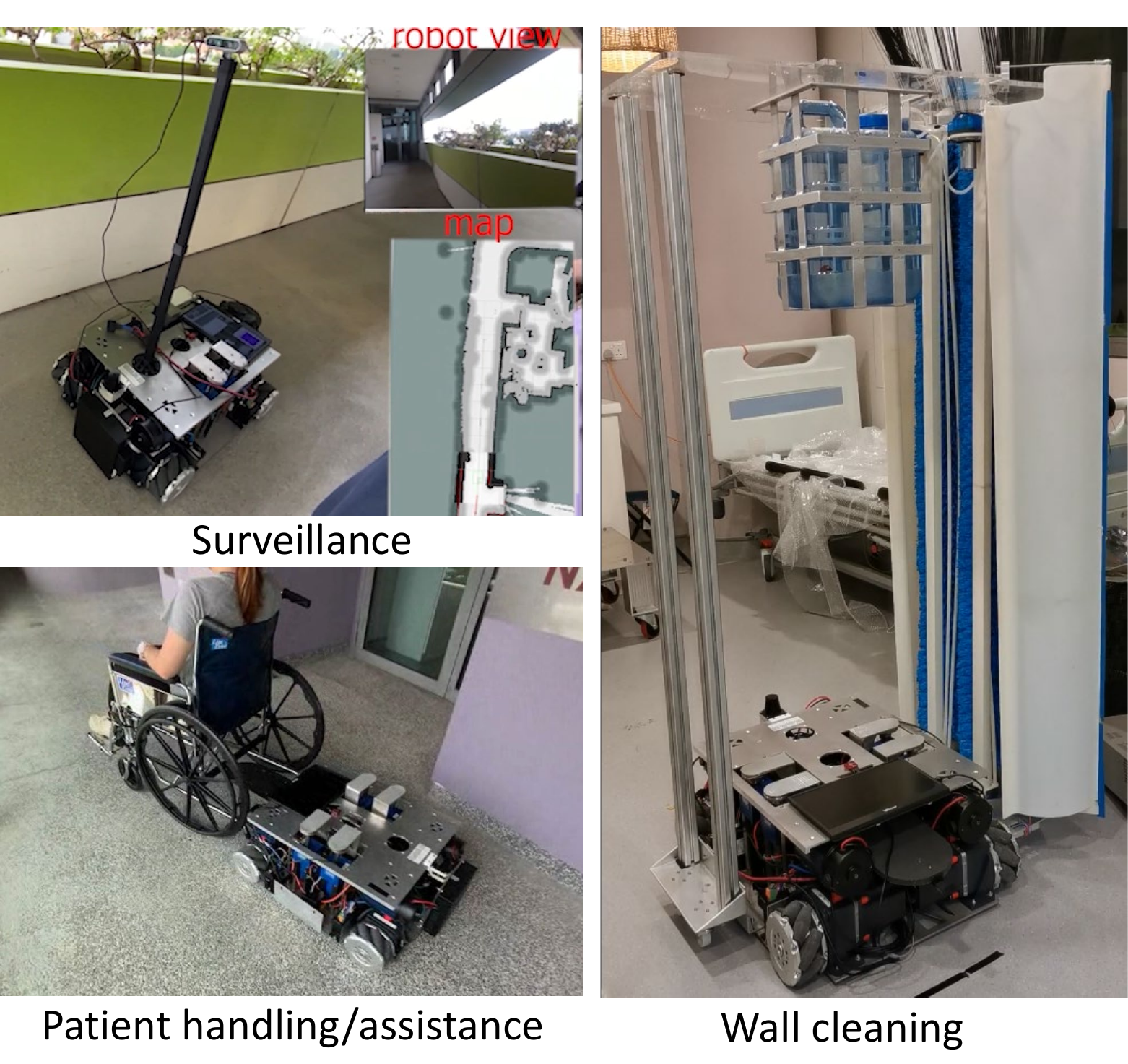}
	\caption{Three possible use cases other than logistics}
	\label{fig:MoreUse}
\end{figure}
To further enlarge the pool of payloads that can be transported using \textit{WaspL}, the robot is designed using nested-reconfigurable robot as classified in \cite{tan2020framework}. The intra-reconfigurability shown in Fig. \ref{fig:modesusecase} is the three different modes aim to move the typical trolleys mentioned in Section \ref{sec:DP}. Whereas the inter-reconfigurability of the robot is designed for other purposes such as trolleys that have larger base as demonstrated in Fig. \ref{fig:modesnested}.

Moreover, other than being used as a logistic robot, \textit{WaspL} is also identified to have some other use cases, such as, wall cleaning, patient handling and surveillance. Fig. \ref{fig:MoreUse} shows different modules that can be connected to \textit{WaspL} and perform its own tasks.

The wall cleaning module consists of a disinfectant container, three brushes and two motorized pumps to clean the wall up to 1.8 meters tall which is the most frequently touched area. Surveillance module is just made up of a camera that is 1.3 meters above the ground, which can be used to detect suspicious items, mask-wearing or physical distancing during this COVID-19 pandemic. The implementation of these three modules on the same \textit{WaspL} platform can be found in the supplementary video.\footnote{https://www.dropbox.com/sh/yn8ty9k6aezbwtj/AAB-v9jptSiPrucTr1Zl7AQIa?dl=0}

%%%%%%%%%%%%%%%%%%%%%%%%%%%%%%%%%%%%%%%%%%%%%%%%%%%%%%%%%%%%%%%%%%%%%%%%%%%%%%%%
%%%%%%%%%%%%%%%%%%%%%%%%%%%%%%%%%%%%%%%%%%%%%%%%%%%%%%%%%%%%%%%%%%%%%%%%%%%%%%%%
\section{Experimental Validation}\label{sec:experi}
\begin{figure}[t]
    \centering
    \includegraphics[width=0.99\linewidth]{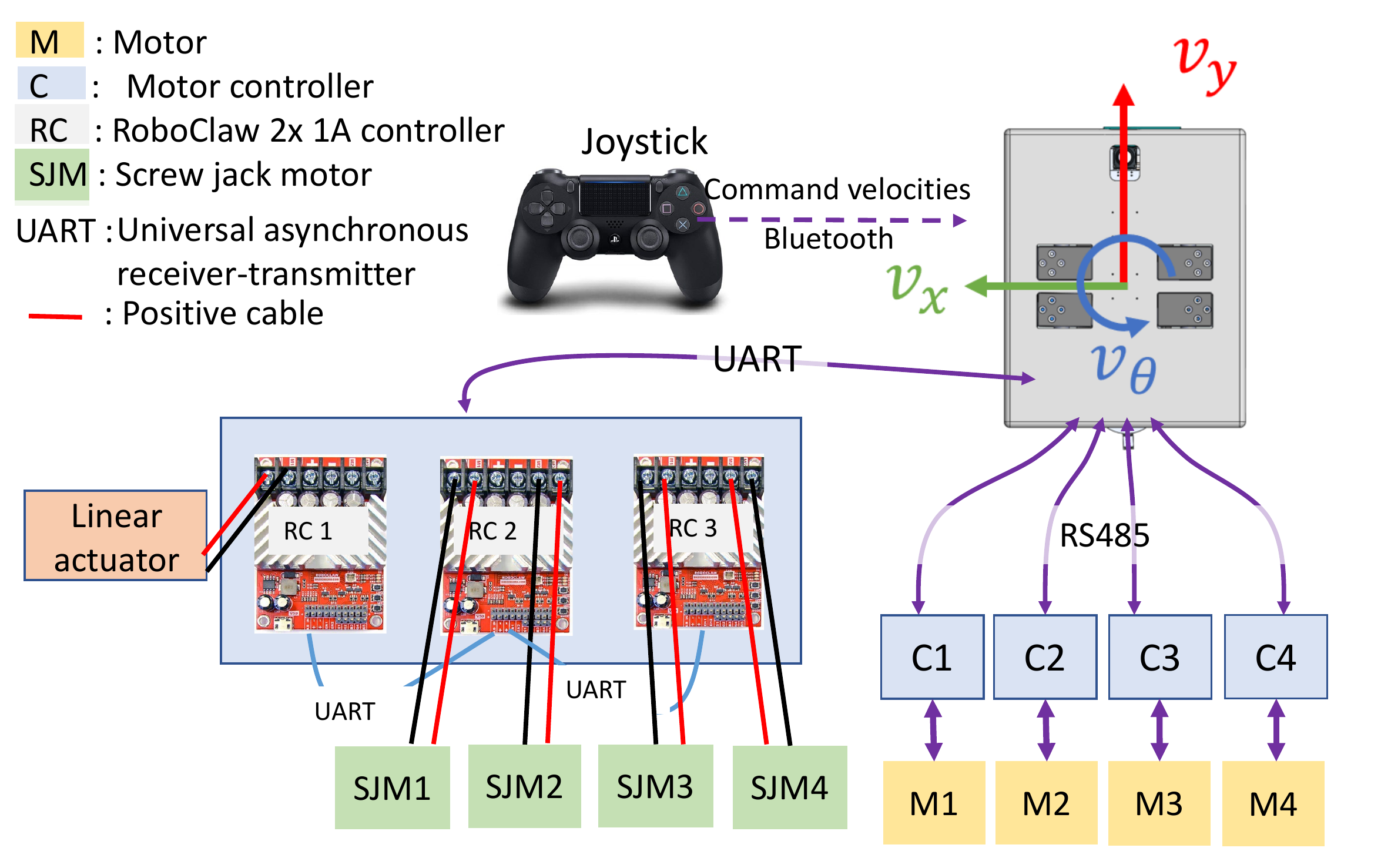}
    \caption{Three possible use cases other than logistics}
    \label{fig:system}
\end{figure}

To validate the multi-functional concept of the \textit{WaspL} system, several experiments were performed using the robot and mock-up trolley bases which are identical to the real trolleys used in the hospital settings at Changi General Hospital (CGH). The system architecture of the experimental setup is shown in Fig. \ref{fig:system}.

\begin{figure}
    \centering
    \includegraphics[width=0.99\linewidth]{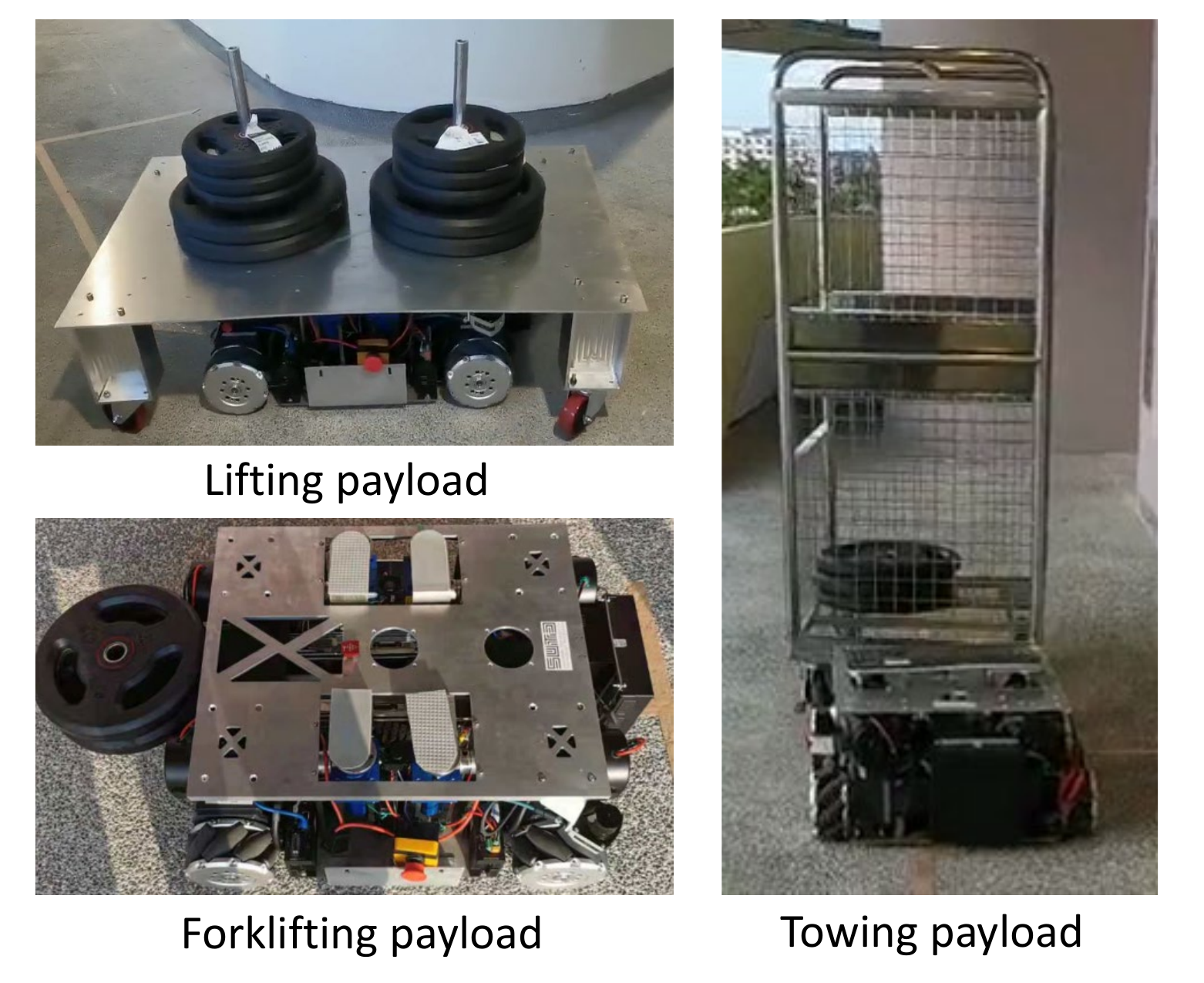}
    \caption{Testing scenarios with three different payloads}
    \label{fig:testing}
\end{figure}

The mobility test of the robot is divided into two parts. In the first test, the robot is given a sequence of velocities through a wireless joystick. The command velocities and the real velocities calculated from the wheel's wheel odometry are recorded and compared. The velocity commands include linear motion on two perpendicular axes of the robot and angular motion. In the second test, the combination of linear and angular velocities is sent out from the joystick for the robot to turn around the corridor's corner. Each of the experiments is tested under four scenarios, as shown in Fig. \ref{fig:testing}, which are a) robot without any payload b) robot with 170kgs lifting payload, c) robot with 20kgs forklift payload and d) robot with 100kgs towing payload. The mobility test's goal is to prove the design of the robot fits all the used cases as proposed and different types of payloads on the robot do not affect the mobility of it.

\begin{figure}
	\centering
	\includegraphics[width=0.99\linewidth]{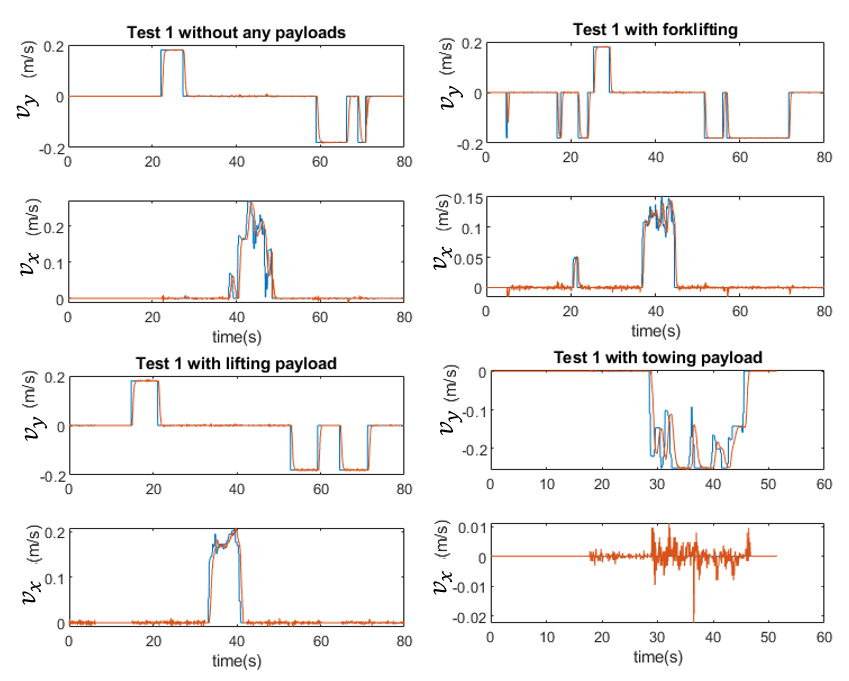}
	\caption{Velocity plots of Test-1 under four scenarios}
	\label{fig:test1result}
\end{figure}

\begin{figure}
	\centering
	\includegraphics[width=0.99\linewidth]{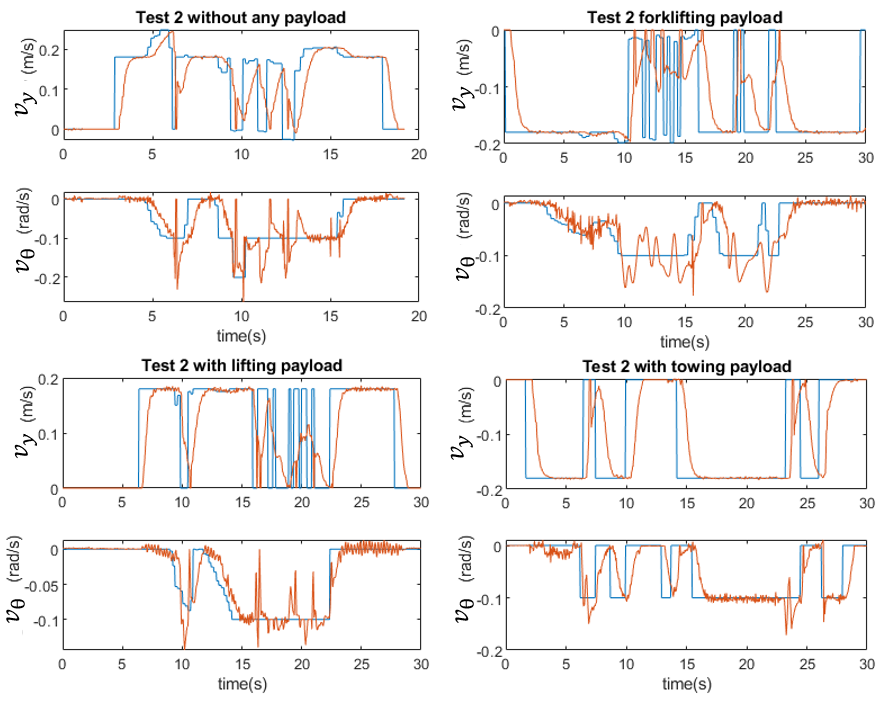}
	\caption{Velocity plots of Test-2 under four scenarios}
	\label{fig:test2result}
\end{figure}

The result of the two tests are plotted separately in Fig. \ref{fig:test1result} and Fig. \ref{fig:test2result}, and the corresponding root means square errors (RMSE) of the velocities are listed in TABLE. \ref{tab:rmse1}. Fig. \ref{fig:test1result} depicts the outcome of Test-1, the $Y$-axis of the plots are the linear velocities of the robot along $v_y$ and $v_x$ as denoted in Fig. \ref{fig:system}. The robot is commended to both $v_y$ and $v_x$ for all the payload types except for the towing payloads. This is because the trolley used for towing test is only equipped with two Swedish caster wheels, which is also identical to the real laundry trolley in the hospital. Hence, it is impossible to move the robot sideways while towing, as it conflicts the trolley's mobility. Therefore, only $v_y$ is given to the robot while it is under the towing scenario. 

Fig. \ref{fig:test2result} shows the results of Test-2 for the robot's velocity commanded and acheived while moving along a curved path. The command velocities given to the robot are combination of $v_y$ and $v_\theta$. Both of the figures speak to the robot's mobility and its ability to follow a given velocity. Table \ref{tab:rmse1} shows that the change of different transportation methods does not significantly influence the platform's control, which further proves that the \textit{WaspL} is adaptive to different types of payloads and logistic modes.

\begin{table}
\caption{Root mean square error in velocities}
 \centering
\begin{tabular}{l|c c c}
Payload type & RMSE $v_{y}$ & RMSE $v_{x}$ & RMSE $v_{\theta}$\\
            & test 1 & test 1 & test 2\\
\hline
Without payload     & 0.0309 & 0.0223 & 0.0383 \\
Lifting payload    & 0.0321 & 0.0309 & 0.0189\\
Forklifting payload     & 0.0354 & 0.0109 & 0.0332\\
Towing payload & 0.0380 & 0.0020 & 0.0376
\end{tabular}

\label{tab:rmse1}
\end{table}

%%%%%%%%%%%%%%%%%%%%%%%%%%%%%%%%%%%%%%%%%%%%%%%%%%%%%%%%%%%%%%%%%%%%%%%%%%%%%%%%
\section{Conclusions}\label{sec:conclusion}
In this paper, we presented the design process of a novel logistic robot named \textit{WaspL}, which can handle different types of payloads using three different mechanisms. This platform's design process involves a case study of the requirement from a real hospital, which leads to a design that combines lifting, towing and forklifting mechanism in a single platform using the design principles. The multi-functionality design makes it possible to handle trolleys with various heights of ground clearances while keeping the overall dimension and footprint minimum. Two experiments were performed to prove the mobility of the platform under different types of payload. The results show that the robot can follow the command velocity accurately under all three use cases. Future research will focus on the control method under the towing mode and the autonomy of the robot.

%%%%%%%%%%%%%%%%%%%%%%%%%%%%%%%%%%%%%%%%%%%%%%%%%%%%%%%%%%%%%%%%%%%%%%%%%%%%%%%%
\section*{Acknowledgements}
This research is supported by the National Robotics Programme under its Robotics Enabling Capabilities and Technologies (Funding Agency Project No. 192 25 00051), National Robotics Programme under its Robot Domain Specific (Funding Agency Project No. 192 22 00058) and administered by the Agency for Science, Technology and Research. 
%%%%%%%%%%%%%%%%%%%%%%%%%%%%%%%%%%%%%%%%%%%%%%%%%%%%%%%%%%%%%%%%%%%%%%%%%%%%%%%%

\bibliographystyle{IEEEtran}
\bibliography{IEEEabrv,panbib}

% Generated by IEEEtran.bst, version: 1.14 (2015/08/26)
\begin{thebibliography}{10}
\providecommand{\url}[1]{#1}
\csname url@samestyle\endcsname
\providecommand{\newblock}{\relax}
\providecommand{\bibinfo}[2]{#2}
\providecommand{\BIBentrySTDinterwordspacing}{\spaceskip=0pt\relax}
\providecommand{\BIBentryALTinterwordstretchfactor}{4}
\providecommand{\BIBentryALTinterwordspacing}{\spaceskip=\fontdimen2\font plus
\BIBentryALTinterwordstretchfactor\fontdimen3\font minus
  \fontdimen4\font\relax}
\providecommand{\BIBforeignlanguage}[2]{{%
\expandafter\ifx\csname l@#1\endcsname\relax
\typeout{** WARNING: IEEEtran.bst: No hyphenation pattern has been}%
\typeout{** loaded for the language `#1'. Using the pattern for}%
\typeout{** the default language instead.}%
\else
\language=\csname l@#1\endcsname
\fi
#2}}
\providecommand{\BIBdecl}{\relax}
\BIBdecl

\bibitem{IQLogistic}
``{Warehouse Automation Market },''
  \url{https://www.thelogisticsiq.com/research/warehouse-automation-market/ },
  accessed: 2021-25-02.

\bibitem{report}
``{Executive Summary WR 2020 Service Robots },'' \url{shorturl.at/iUVZ5 },
  accessed: 2021-25-02.

\bibitem{tamantini2021robotic}
C.~Tamantini, F.~S. di~Luzio, F.~Cordella, G.~Pascarella, F.~E. Agro, and
  L.~Zollo, ``A robotic health-care assistant for covid-19 emergency: A
  proposed solution for logistics and disinfection in a hospital environment,''
  \emph{IEEE Robotics \& Automation Magazine}, 2021.

\bibitem{kriegel2021requirements}
J.~Kriegel, C.~Rissbacher, L.~Reckwitz, and L.~Tuttle-Weidinger, ``The
  requirements and applications of autonomous mobile robotics (amr) in
  hospitals from the perspective of nursing officers,'' \emph{International
  Journal of Healthcare Management}, pp. 1--7, 2021.

\bibitem{SwissLog}
``{Swisslog-healthcare},''
  \url{https://www.swisslog-healthcare.com/en-sg/solutions/transport },
  accessed: 2021-25-02.

\bibitem{Tug}
``{Aethon-TUG},'' \url{https://aethon.com/mobile-robots-for-healthcare/},
  accessed: 2021-25-02.

\bibitem{mir_robot}
``{MiR Robot},'' \url{https://www.healthcaredenmark.dk/danish-solutions/mir/},
  accessed: 2021-25-02.

\bibitem{fadzil2013development}
C.~M. Fadzil, ``Development of automated-guided vehicle for 300kg trolley
  towing application,'' 2013.

\bibitem{horan2011oztug}
B.~Horan, Z.~Najdovski, T.~Black, S.~Nahavandi, and P.~Crothers, ``{OzTug}
  mobile robot for manufacturing transportation,'' in \emph{2011 IEEE
  International Conference on Systems, Man, and Cybernetics}.\hskip 1em plus
  0.5em minus 0.4em\relax IEEE, 2011, pp. 3554--3560.

\bibitem{garibott1996robolift}
G.~Garibott, S.~Masciangelo, M.~Ilic, and P.~Bassino, ``Robolift: a vision
  guided autonomous fork-lift for pallet handling,'' in \emph{Proceedings of
  IEEE/RSJ International Conference on Intelligent Robots and Systems.
  IROS'96}, vol.~2.\hskip 1em plus 0.5em minus 0.4em\relax IEEE, 1996, pp.
  656--663.

\bibitem{wang2010innovative}
J.-Y. Wang, J.-S. Zhao, F.-L. Chu, and Z.-J. Feng, ``Innovative design of the
  lifting mechanisms for forklift trucks,'' \emph{Mechanism and Machine
  Theory}, vol.~45, no.~12, pp. 1892--1896, 2010.

\bibitem{siddiqi2008modeling}
A.~Siddiqi and O.~L. de~Weck, ``Modeling methods and conceptual design
  principles for reconfigurable systems,'' \emph{Journal of Mechanical Design},
  vol. 130, no.~10, 2008.

\bibitem{tan2020framework}
N.~Tan, A.~A. Hayat, M.~R. Elara, and K.~L. Wood, ``A framework for taxonomy
  and evaluation of self-reconfigurable robotic systems,'' \emph{IEEE Access},
  vol.~8, pp. 13\,969--13\,986, 2020.

\bibitem{singh2009innovations}
V.~Singh, S.~M. Skiles, J.~E. Krager, K.~L. Wood, D.~Jensen, and
  R.~Sierakowski, ``Innovations in design through transformation: a fundamental
  study of transformation principles,'' \emph{Journal of Mechanical Design},
  vol. 131, no.~8, 2009.

\bibitem{weaver2010transformation}
J.~Weaver, K.~Wood, R.~Crawford, and D.~Jensen, ``Transformation design theory:
  a meta-analogical framework,'' \emph{Journal of Computing and Information
  Science in Engineering}, vol.~10, no.~3, 2010.

\bibitem{Xnergy}
``{Xnergy Wireless universal charging},'' \url{https://www.xnergytech.com/},
  accessed: 2021-28-02.

\bibitem{MSA_AGV}
``{MSA AGV},'' \url{https://www.msarobotics.com.au/agvs}, accessed: 2021-25-02.

\bibitem{Elfin_OKGAV}
``{OKGAV Elfin},''
  \url{http://www.okagv.com/Warehouse_logistics_AGV_robots_15452400.html},
  accessed: 2021-25-02.

\bibitem{Uwant}
``{UWANT},'' \url{ }, accessed: 2021-25-02.

\bibitem{OmniTurtle}
``{OmniTurtle HTT Logistics},''
  \url{http://www.hangfa.com/EN/omniturtle/htt.html }, accessed: 2021-25-02.

\bibitem{MicroLomy}
``{MircoLomay Logistic Vehicle},''
  \url{https://www.i-so.com.cn/en/MagneticAGV/84.html }, accessed: 2021-25-02.

\bibitem{KUKA_Robot}
``{KUKA robotic KMP1500 },'' \url{https://cutt.ly/2lHRKuO }, accessed:
  2021-25-02.

\end{thebibliography}

\end{document}